\documentclass[letterpaper, 10 pt, conference]{ieeeconf}  

\IEEEoverridecommandlockouts                              

\usepackage{graphics}
\usepackage{amsmath} 
\usepackage{amssymb}  
\usepackage{ifpdf}
\usepackage{cite}
\usepackage[pdftex]{graphicx}
\usepackage{soul}
\usepackage{makecell}
\usepackage{bm}
\usepackage{color}
\usepackage{algorithmic}
\usepackage{array}
\usepackage{url}
\usepackage{hyperref}
\usepackage{multirow}
\usepackage{siunitx}
\usepackage{subfigure}
\usepackage{placeins}
\usepackage{caption}
\usepackage{booktabs}
\usepackage{stfloats}
\usepackage{comment}
\usepackage{wasysym}
\usepackage{amssymb}
\usepackage{pifont}
\usepackage{wrapfig}
\graphicspath{{figures/}}
\DeclareGraphicsExtensions{.pdf,.jpeg,.png,.jpg}

\title{\LARGE \bf
Give me scissors: Collision-Free Dual-Arm Surgical Assistive Robot for Instrument Delivery
}

\author{Xuejin Luo, Shiquan Sun, Runshi Zhang, Ruizhi Zhang, Junchen Wang$^\ast$
\thanks{This work was supported in part by the Natural Science Foundation of China under Grant 62573022, Grant U22A2051; and in part by the Natural Science Foundation of Beijing Municipality under Grant L232037. (\textit{Corresponding author: Junchen Wang})}
\thanks{X. Luo, S. Sun, R. Zhang, R. Zhang, and J. Wang are with the School of Mechanical Engineering and Automation, Beihang University, Beijing, China.
        {\tt\small wangjunchen@buaa.edu.cn}}}

\begin{document}

\maketitle
\thispagestyle{empty}
\pagestyle{empty}

\begin{abstract}

During surgery, scrub nurses are required to frequently deliver surgical instruments to surgeons, which can lead to physical fatigue and decreased focus.
Robotic scrub nurses provide a promising solution that can replace repetitive tasks and enhance efficiency.
Existing research on robotic scrub nurses relies on predefined paths for instrument delivery, which limits their generalizability and poses safety risks in dynamic environments.
To address these challenges, we present a collision-free dual-arm surgical assistive robot capable of performing instrument delivery.
A vision-language model is utilized to automatically generate the robot's grasping and delivery trajectories in a zero-shot manner based on surgeons' instructions.
A real-time obstacle minimum distance perception method is proposed and integrated into a unified quadratic programming framework.
This framework ensures reactive obstacle avoidance and self-collision prevention during the dual-arm robot's autonomous movement in dynamic environments.
Extensive experimental validations demonstrate that the proposed robotic system achieves an 83.33\% success rate in surgical instrument delivery while maintaining smooth, collision-free movement throughout all trials.
The project page and source code are available at https://give-me-scissors.github.io/.

\end{abstract}

\section{INTRODUCTION}

The scrub nurse is a vital member of the surgical team, primarily responsible for assisting the surgeon with tasks such as delivering surgical instruments and managing retractors.
However, this work is highly mechanized and repetitive, which can lead to physical fatigue and potential errors among scrub nurses~\cite{ng2024multimodal}.
Additionally, in units understaffed with nurses, the lack of a scrub nurse leads to a significant decrease in the efficiency of the surgical team, potentially resulting in more serious safety issues~\cite{jacob2011gesture}.
New technological methods, such as robotic-assisted surgery and automation tools, present a promising solution to these challenges~\cite{xu2015robotic}.
These advanced technologies not only alleviate the burden on scrub nurses but also enhance efficiency of surgical team~\cite{ng2025large}.

Researchers have developed various surgical assistive robots to function as scrub nurses and carry out the highly mechanized task of surgical instrument delivery.
A robotic scrub nurse was designed for safe human-robot collaboration in the operating room~\cite{jacob2013surgical}.
Surgeons can request surgical instruments through speech commands and hand gestures.
The dual-arm robotic system was proposed for surgical instruments transferring tasks in~\cite{wu2019coordinated}.
A deep learning-based multi-modal robotic framework was presented in~\cite{ng2024multimodal}, which can work with surgeons via speech and image inputs.
However, the categories of surgical instruments and the delivery paths must be predefined in these approaches, which limits their autonomy and generalization.
Furthermore, the absence of real-time collision avoidance capabilities poses safety challenges in dynamic and complex surgical environment~\cite{xian2023semi}.

\begin{figure}[t] 
\centering
\includegraphics[width=.9\linewidth]{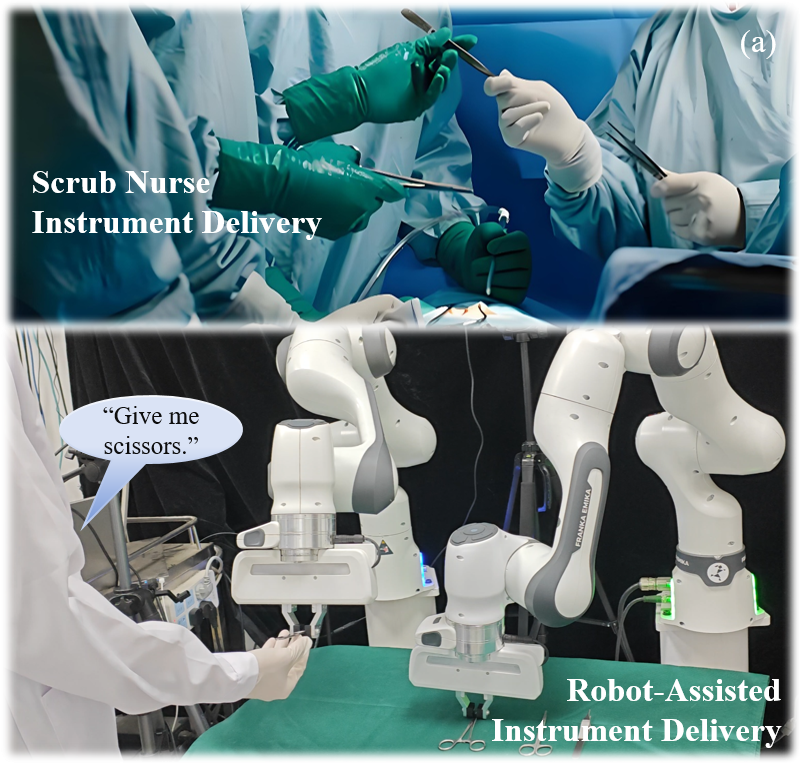}
\caption{
        Surgical instruments transfer process.
        (a) Scrub nurse instrument delivery.
        (b) Robot-assisted instrument delivery.
}
\label{background}
\end{figure}

In recent years, Vision Language Models (VLMs) have been widely used in robotic autonomy tasks~\cite{hu2023toward}.
They have demonstrated remarkable performance in scene perception and motion planning.
VLM is integrated into the robotic planning system in~\cite{Gao2024ICRA}, enabling the combination of language instructions and image inputs to generate high-level task plans.
VoxPoser~\cite{huang2023voxposer} applies VLM to obtain the 3D value map used within a planning framework to synthesize closed-loop robot trajectories.
ReKep~\cite{CoRL2024Huang} utilizes VLM to associate objects in the environment with keypoints, generating constraints relevant to navigation tasks.
T-Rex~\cite{chen2025t} proposes a VLM-based task-adaptive spatial representation extraction framework for robotic manipulation.
Benefiting from VLMs, the robots in these studies demonstrate capabilities in scene semantic understanding and action grounding.
In zero-shot tasks, these robots achieved autonomous planning and movement, indicating the excellent generalization abilities.
However, VLM-based autonomous planning is an emerging field in robotic-assisted surgery.
In the surgical assistive robotic system, leveraging VLMs for motion planning offers distinctive advantages.
On one hand, their multi-modal capabilities enable robots to intuitively comprehend the surgeon's instructions.
On the other hand, their autonomous planning ability allows surgical assistive robots to adapt to the dynamic intraoperative environment.

Autonomous motion of robots imposes high demands on system safety~\cite{luo2025tase, WRC2023luo}.
Ensuring collision-free operation during robotic-assisted surgery is particularly challenging~\cite{xian2025task}.
Traditional sampling-based global planners can generate collision-free paths offline~\cite{TRO2024Laha}.
However, high computational cost makes them unsuitable for dynamic and unstructured environments~\cite{RAL2021Mikhail}.
Several methods for real-time obstacle avoidance have been reported.
A reactive collision avoidance method was presented in~\cite{IJRR2024Mikhail}, which generated collision-free motion in joint space.
A high-order control barrier functions (CBFs) framework for collision avoidance among convex primitives was proposed in~\cite{Wei2025TCST}.
ToMPC~\cite{jia2025tompc} introduced a task-oriented Model Predictive Control (MPC) framework for safe and efficient robotic manipulation in open workspaces.
Nevertheless, these works rely on visual markers and sensors, which limits their application in unstructured environments.
In addition to obstacle avoidance, dual-arm systems must also consider self-collision avoidance due to the overlap of their workspaces~\cite{TASE2024Zhang}.
A real-time dual-arm self-collision avoidance method was proposed in~\cite{luo2025tase}.
Building on these previous efforts regarding obstacle and self-collision avoidance, the dual-arm surgical assistive robot faces unique requirements.
It must detect obstacles in an unstructured dynamic environment without the use of markers, and perform reactive obstacle avoidance and self-collision avoidance simultaneously.

\begin{figure*}[thpb] 
\centering
\includegraphics[width=\textwidth]{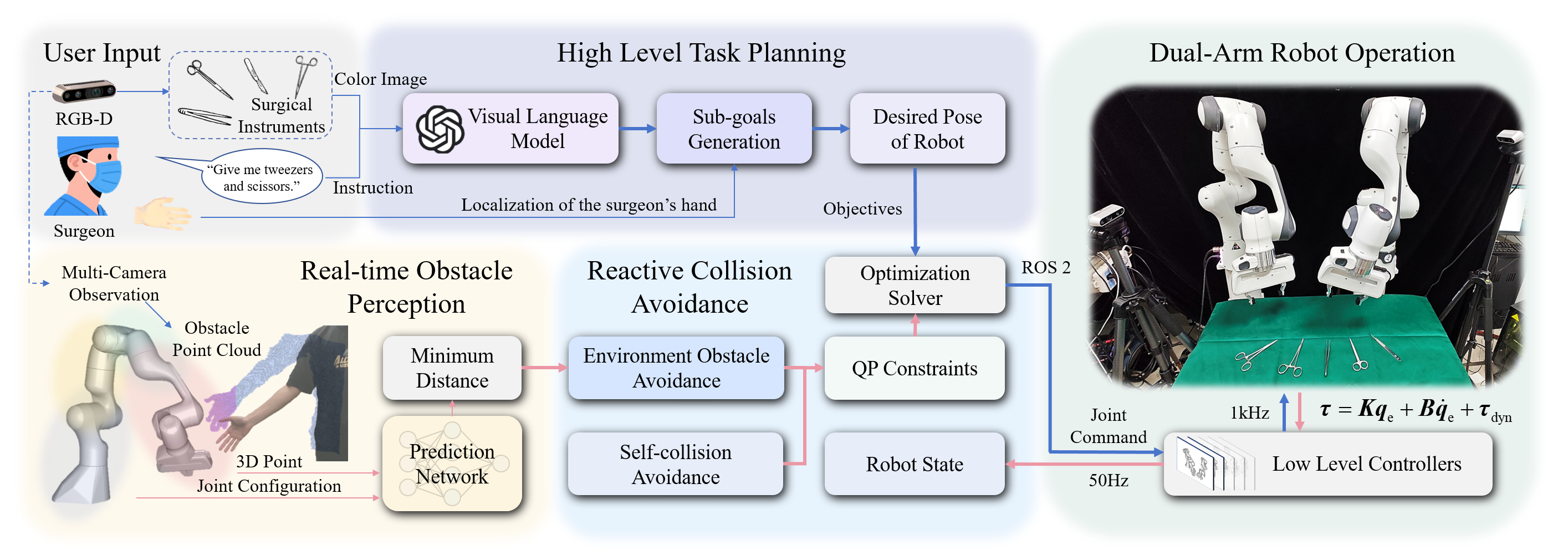}
\caption{
        The pipeline of the collision-free dual-arm surgical assistive robot for instrument delivery.
        The robot system receives multi-modal inputs from the physical world (i.e., surgeon instruction, color image, depth data).
        The real-time obstacle perception module computes the minimum distance between robot and environmental obstacles.
        The QP framework is built upon the minimum distance, ensuring the dual-arm robot's collision-free operation.
        The high-level task planning utilizes VLM to generate the desired motion objectives for the QP framework.
}
\label{pipeline}
\end{figure*}

In this work, we present a collision-free dual-arm surgical assistive robot for instrument delivery.
It utilizes VLM to achieve scene semantic understanding, autonomously planning the robot's grasping and delivery paths based on multi-modal inputs, including surgeon's instructions and visual features of surgical instruments. 
The entire \textbf{\textit{zero-shot}} process does not need fine-tuning or predefined operations.
During autonomous motion, the robot maintains high real-time perception of the nearest obstacles, without the need for visual markers or prior environmental modeling.
A unified quadratic programming (QP) framework is utilized to achieve reactive obstacle avoidance and self-collision avoidance based on minimum distance perception.
Extensive real-world experimental validation in collision avoidance and surgical instrument delivery has confirmed the system's robustness and safety.
The main contributions are as follows:
\begin{itemize}
\item A dual-arm surgical assistive robot for instrument delivery is developed. It utilizes VLM to automatically generate the robot's grasping and delivery motion based on surgeon's instructions.
\item A unified real-time QP framework is proposed to achieve dual-arm robot reactive obstacle avoidance and self-collision avoidance simultaneously during the autonomous movement.
\item The proposed robotic system achieved a success rate of 83.33\% in real-world instrument delivery experiments, with no collisions occurring, thereby demonstrating its effectiveness and safety.
\end{itemize}

\section{Method}

The overview of the dual-arm surgical assistive robot is illustrated in Fig.~\ref{pipeline}.
A real-time obstacle perception method is proposed, predicting the distance from the robotic arm links to obstacles.
The minimum distance is used to construct the nonlinear constraint in the QP framework to ensure real-time collision avoidance.
The proposed QP framework functions as a safety filter, achieving motion objectives while ensuring that the robot satisfies constraints for obstacle avoidance, self-collision avoidance, and joint limits.
The VLM generates task level sub-goals by interpreting multi-modal inputs from surgeon's commands and camera observations, providing motion objectives for the QP framework.
The methodology is elaborated as follows:
Section~\ref{MinimumDistancePrediction} introduces the process of real-time obstacle perception. 
Section~\ref{QPframework} describes the architecture of the reactive collision avoidance QP framework. 
Section~\ref{TaskPlanning} presents the task-planning mechanism enabled by the VLM.

\subsection{Real-time Obstacle Perception}
\label{MinimumDistancePrediction}
The real-time perception of environmental obstacles is essential for achieving reactive collision avoidance.
The primary goal is to identify the closest point to the robot and the direction in which the robot should move away.
Let $\mathbb{P}$ denote the set of all point clouds in the environment.
To reduce computational cost, $\mathbb{P}$ is filtered as follows to isolate the point cloud in the robot's vicinity.
For each joint configuration $\boldsymbol{q}$, the occupancy volumes of n links $\mathbb{S}_{\mathrm{r}}(\boldsymbol{q}) = \left\{s_{\mathrm{1}}, s_{\mathrm{2}}, \ldots, s_{\mathrm{n}}\right\} \subset \mathbb{R}^{3}$ are determined through forward kinematics (FK), where $s_{\mathrm{i}} \subset \mathbb{R}^{3}$ represents the three-dimensional volume occupied by the $i$-th link.
To accelerate collision detection, the manipulator's complex geometry is approximated by a capsule model, denoted as $\mathbb{S}_{\mathrm{cap}}(\boldsymbol{q})$:
\begin{align}
\mathbb{S}_{\mathrm{cap}}(\boldsymbol{q}) = \bigcup_{i=1}^{n} \left\{ \mathbb{P} \in \mathbb{R}^3 \mid d(\mathbb{P}, l_i) \leqslant r_{\mathrm{cap}} \right\} 
\label{capsule}
\end{align}
where $l_i$ is the line segment connecting the adjacent joint centers $p_{i1}$ and $p_{i2}$.
$r_{\mathrm{cap}}$ is the radius of the capsule.
$d(\mathbb{P}, l_i)$ denotes the minimum distance from a point to the line segment.
Eq.~\ref{capsule} enables the rapid identification of points that lie inside safety capsules. 
It avoids computing the distance from $\mathbb{P}$ to every point on the complex mesh $\mathbb{S}_{\mathrm{r}}(\boldsymbol{q})$, which have a time complexity of $O(n^2)$. 

\begin{figure}[thpb] 
\centering
\includegraphics[width=\linewidth]{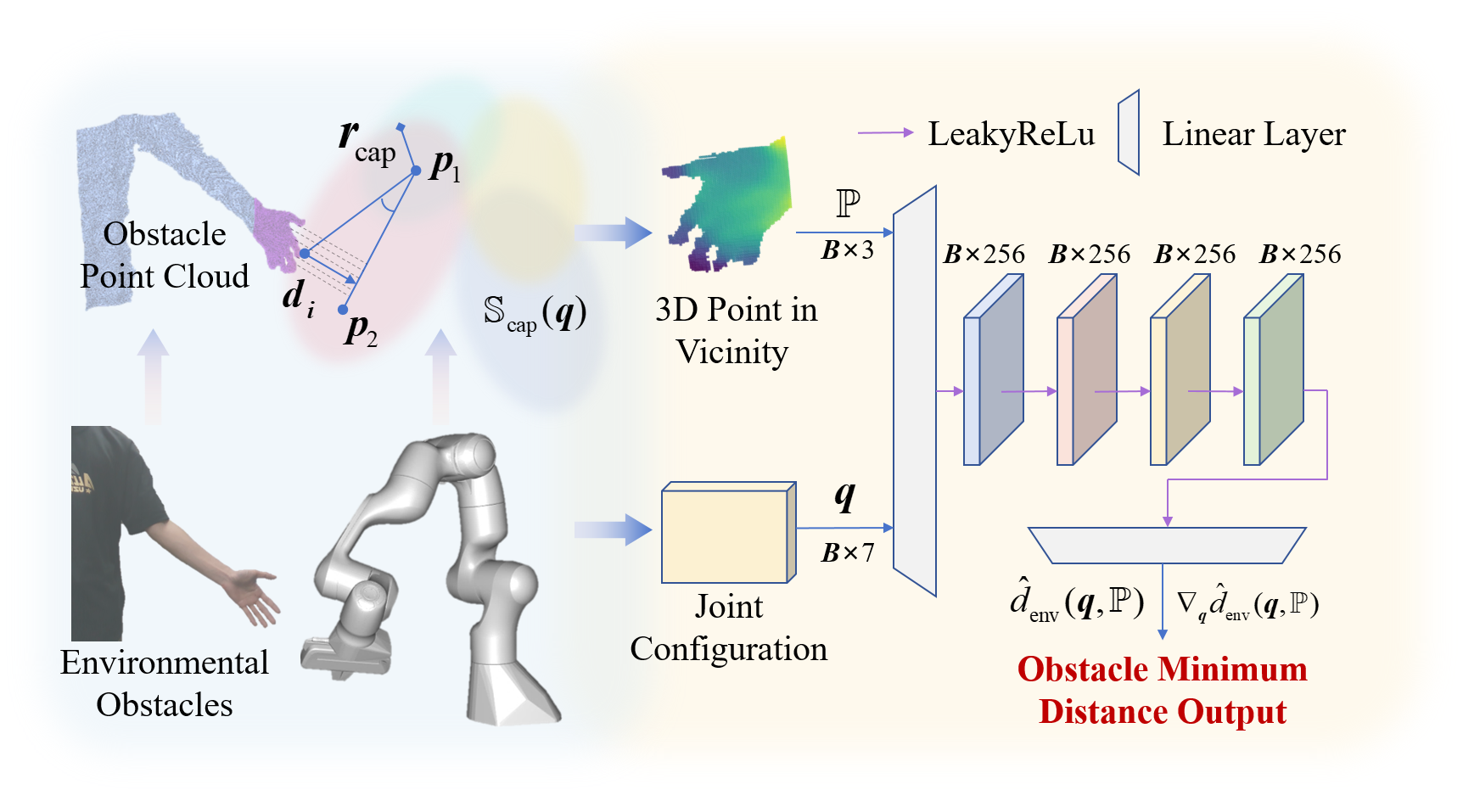}
\caption{
        Real-time obstacle perception process.
        The robot joint configuration and point cloud of obstacles are taken as input.
        Output is the minimum distance between robot and obstacle.
}
\label{EnvDistancePrediction}
\end{figure}

The set $\mathbb{P}$ within $\mathbb{S}_{\mathrm{cap}}(\boldsymbol{q})$ contains a substantial number of points belonging to the robot. 
To prevent self-collision interference, these points must be filtered. 
The filtering process is performed by first generating the 2D robot mask via image segmentation~\cite{sundaralingam2023curobo}. 
Then the mask is mapped to the synchronized depth map to extract the robot's point cloud. 

Then the robot vicinity $\mathbb{S}_{\mathrm{v}} \subset \mathbb{R}^{3}$ is calculated as follows
\begin{align}
\mathbb{S}_{\mathrm{v}}(\boldsymbol{q}) = \left\{ \mathbb{S}_{\mathrm{cap}}(\boldsymbol{q}) \cap \bar{\mathbb{S}}_{\mathrm{r}}(\boldsymbol{q}) \right\} 
\label{vicinity}
\end{align}
Let $\mathbb{P}_{\mathrm{v}} \subset \mathbb{R}^{3}$ be the set of environmental point cloud in $\mathbb{S}_{\mathrm{v}}(\boldsymbol{q})$.
$\bar{\mathbb{S}}_{\mathrm{r}}(\boldsymbol{q})$ denotes the complement of $\mathbb{S}_{\mathrm{r}}(\boldsymbol{q})$, representing the points that do not belong to the robot.
The closest point $\mathbb{P}_{\mathrm{min}} \in \mathbb{P}_{\mathrm{v}}$ to $\mathbb{S}_{\mathrm{r}}(\boldsymbol{q})$ and the corresponding minimum distance $d_{\mathrm{min}}$ are calculated as
\begin{align}
\begin{split}
\mathbb{P}_{\mathrm{min}} = &\underset{\mathbb{P}_{i} \in \mathbb{P}_{\mathrm{v}}}{\text{argmin}} \left\lVert \mathbb{P}_{i} - \mathbb{S}_{\mathrm{r}}(\boldsymbol{q})\right\rVert ^2 \\
d_{\mathrm{min}} &= \left\lVert \mathbb{P}_{\mathrm{min}} - \mathbb{S}_{\mathrm{r}}(\boldsymbol{q})\right\rVert
\end{split}
\label{ClosestPoint}
\end{align}
where $\mathbb{P}_{i}$ is the $i$-th point in $\mathbb{P}_{\mathrm{v}}$.
To avoid high computational costs, a distance prediction neural network $\hat{d}_{\mathrm{env}}{(\boldsymbol{q}, \mathbb{P})}$ is proposed to approximate the $d_{\mathrm{min}}$.
The network architecture is illustrated in Fig.~\ref{EnvDistancePrediction}.
The inputs include $\mathbb{P}_{\mathrm{v}}$ and the current robot joint configuration $\boldsymbol{q}$.
Linear layers with high-dimensional features are employed to capture the implicit relationships between the obstacle points and the robot.
Nonlinear activation functions are utilized to model the complex interactions.
The output consists of the minimum distance $\hat{d}_{\mathrm{env}}{(\boldsymbol{q}, \mathbb{P})}$ and gradient obtained through backpropagation.

\subsection{Reactive Collision Avoidance Framework}
\label{QPframework}
A QP framework is utilized to achieve reactive collision avoidance.
The QP framework is formulated as follows:
\begin{align}
\begin{split}
\Delta \boldsymbol{q}_{\mathrm{de}} = \underset{ \Delta \boldsymbol{q} }{\text{argmin}} 
\frac{1}{2} \alpha 
\underbrace{
\left\lVert \Delta \boldsymbol{q} - \boldsymbol{J^\dagger}(\boldsymbol{q}_{\mathrm{c}})\boldsymbol{v}_{\mathrm{de}} \right\rVert  ^{2} 
}_{\mathrm{A}}
\\
\quad \quad + \frac{1}{2} \beta 
\underbrace{
\left\lVert \Delta \boldsymbol{q} + \boldsymbol{q}_{\mathrm{c}} - \boldsymbol{q}_{\mathrm{de}} \right\rVert  ^{2} 
}_{\mathrm{B}}
+ \underbrace{
\Delta \boldsymbol{q}^{T} 
\boldsymbol{Q} 
\Delta \boldsymbol{q}
}_{\mathrm{C}}
\end{split}
\label{QPobjectivs}
\\
\begin{split}
\mathrm{s.t.} \quad 
\left\{
\begin{array}{l}
\boldsymbol{q} = \boldsymbol{q}_{\mathrm{c}} + \Delta \boldsymbol{q}
\\
\ln \left( \frac{\hat{d}_{\mathrm{env}}{(\boldsymbol{q}, \mathbb{P})}}{\lambda } \right) + \Delta \boldsymbol{q} \cdot \nabla_{\boldsymbol{q}} \hat{d}_{\mathrm{env}}{(\boldsymbol{q}, \mathbb{P})}  \geq  0
\\
\ln \left( \frac{\hat{d}_{\mathrm{self}}(\boldsymbol{q})}{\mu } \right) + \Delta \boldsymbol{q} \cdot \nabla_{\boldsymbol{q}} \hat{d}_{\mathrm{self}}(\boldsymbol{q})  \geq  0
\\
\boldsymbol{q}^{\mathrm{min}} < \boldsymbol{q} < \boldsymbol{q}^{\mathrm{max}}
\\
\zeta^{\mathrm{min}} < \Delta \boldsymbol{q} < \zeta^{\mathrm{max}}
\end{array}
\right.
\end{split}
\label{QPconstraints}
\end{align}
where $\Delta \boldsymbol{q} = \left[ \Delta \boldsymbol{q}^{\mathrm{L}}, \Delta \boldsymbol{q}^{\mathrm{R}} \right]^{T}$ represents the joint-space increments of left and right arm, respectively.
In Eq.~\eqref{QPobjectivs}, the goal is to optimize the desired joint-space $\Delta \boldsymbol{q}_{\mathrm{de}}$ to satisfy dual-arm robot motion tasks while maintaining safety. 
\subsubsection{Cartesian Velocity Objective}
Term A in Eq.~\eqref{QPobjectivs} denotes the optimization objective for dual-arm cartesian velocity.
$\boldsymbol{J^\dagger}(\boldsymbol{q}_{\mathrm{c}})$ is the pseudo-inverse of the Jacobian matrix at current joint configurations $\boldsymbol{q}_{\mathrm{c}}$.
$\boldsymbol{v}_{\mathrm{de}}$ is desired Cartesian velocity.
It is transformed into joint-space increments via $\boldsymbol{J^\dagger}(\boldsymbol{q}_{\mathrm{c}})$, ensuring $\Delta \boldsymbol{q}$ is optimized to approach dual-arm desired Cartesian velocity.
$\alpha$ is a positive weight factor.

\subsubsection{Reference Joint Objective}
Term B in Eq.~\eqref{QPobjectivs} represents the optimization objective for dual-arm robot desired joint configurations.
$\Delta \boldsymbol{q}$ is optimized to narrow the gap between the current joint configurations $\boldsymbol{q}_{\mathrm{c}}$ and the desired joint configurations $\boldsymbol{q}_{\mathrm{de}}$.
$\beta$ is the positive weight factor of term B.
The optimization objectives of Term A and Term B ensure that the motion of the dual-arm robot aligns with the task goals.
Term C is utilized to keep the variation range of $\Delta \boldsymbol{q}$ as small as possible.
$\boldsymbol{Q}$ is the weight matrix.

\subsubsection{Obstacle Collision Constraints}
\label{EnvironmentCollisionConstraints}
The distance prediction neural network $\hat{d}_{\mathrm{env}}{(\boldsymbol{q}, \mathbb{P})}$ proposed in \ref{MinimumDistancePrediction} is utilized to construct the obstacle collision constraints.
$\lambda$ is the safety distance threshold between robot and environment obstacles.
The gradient of $\hat{d}_{\mathrm{env}}{(\boldsymbol{q}, \mathbb{P})}$ with respect to $\boldsymbol{q}$ is given by
\begin{align}
\nabla_{\boldsymbol{q}} \hat{d}_{\mathrm{env}}{(\boldsymbol{q}, \mathbb{P})} 
= \frac{\partial \hat{d}_{\mathrm{env}}{(\boldsymbol{q}, \mathbb{P})}}{ \partial\boldsymbol{q} } 
= \mathcal{B} \left( \hat{d}_{\mathrm{env}}{(\boldsymbol{q}, \mathbb{P})}  \right) 
\label{EnvGradient}
\end{align}
where $\mathcal{B}$ denotes the backpropagation operation.
When $\hat{d}_{\mathrm{env}}{(\boldsymbol{q}, \mathbb{P})}$ is less than $\lambda$, $\ln \left( \frac{\hat{d}_{\mathrm{env}}{(\boldsymbol{q}, \mathbb{P})}}{\lambda } \right)$ will be negative.
Then it forces $\Delta \boldsymbol{q}$ to align with the direction of the gradient $\nabla_{\boldsymbol{q}} \hat{d}_{\mathrm{env}}{(\boldsymbol{q}, \mathbb{P})}$ to satisfy the constraint.
Due to the non-linearity of the logarithm function, the influence of the term $\ln \left( \frac{\hat{d}_{\mathrm{env}}{(\boldsymbol{q}, \mathbb{P})}}{\lambda } \right)$ on $\Delta \boldsymbol{q}$ increases as the robot approaches the environment obstacles.
Conversely, when $\hat{d}_{\mathrm{env}}{(\boldsymbol{q}, \mathbb{P})}$ is greater than $\lambda$, the constraint relaxes, allowing $\Delta \boldsymbol{q}$ to be optimized freely.

\subsubsection{Self-collision Constraints}
In third term of Eq.~\eqref{QPconstraints}, the self-collision constraint is established using the distance prediction neural network $\hat{d}_{\mathrm{self}}(\boldsymbol{q})$.
Inspired by~\cite{luo2025tase}, $\hat{d}_{\mathrm{self}}(\boldsymbol{q})$ is trained to predict the minimum distance between dual arms.
$\mu$ is the corresponding safety distance threshold.
$\nabla_{\boldsymbol{q}} \hat{d}_{\mathrm{self}}(\boldsymbol{q})$ is the gradient of $\hat{d}_{\mathrm{self}}(\boldsymbol{q})$ with respect to $\boldsymbol{q}$, follows a pattern analogous to that in Eq~\eqref{EnvGradient}.
The mechanism of the self-collision avoidance constraint is similar to~\ref{EnvironmentCollisionConstraints}.

The last two lines of Eq.~\eqref{QPconstraints} represent the constraints of joint limits and joint velocity respectively.
$\boldsymbol{q}^{\mathrm{min}}$ and $\boldsymbol{q}^{\mathrm{max}}$ denote the lower and upper bounds of the joints, while $\zeta^{\mathrm{min}}$ and $\zeta^{\mathrm{max}}$ represent the lower and upper limits of joint velocity.
The QP framework~\eqref{QPobjectivs}\eqref{QPconstraints} serves as a real-time safety filter, ensuring collision-free motions of the dual-arm robot in dynamic environment.

\subsection{Task Planning}
\label{TaskPlanning}
High-level task planning provides the desired motion objectives for the QP framework based on the surgeon's commands.
The specific process of high-level task planning is illustrated in Fig.~\ref{pipeline}.
The robot system interprets semantic information from the surgeon and effectively processes multi-modal inputs.
During surgery, the instructions from the surgeon are recorded and converted into text format, denoted as $\boldsymbol{T}$.
Multiple RGB-D cameras are employed to capture color images and depth information of the surgical instruments.
Let the images be denoted as $\boldsymbol{I}$, and the point clouds transformed from depth be represented as $\mathbb{P}_{\mathrm{obj}}$.
We calculate the pixel features $\boldsymbol{X}$ utilizing visual model DINOv2~\cite{oquab2023dinov2}.
Then segmentation model SAM~\cite{Kirillov2023SAM} is applied to get the masks of $n$ objects, denoted as $\boldsymbol{M} = \left\{ \boldsymbol{m}_1, \boldsymbol{m}_2, \ldots, \boldsymbol{m}_n \right\}$.
For each mask $\boldsymbol{m}_{i}$, the 3D keypoint $\boldsymbol{p}_{i}$ is generated as follows
\begin{align}
\boldsymbol{p}_{i} = \underset{ o_{i} \subset \mathbb{R}^{3} }{\text{argmin}} \sum_{j = 1}^{k} \left\{ \left\lVert \mathbb{P}_{j}^{m_{i}} - o_{i} \right\rVert ^2 + \left\lVert \boldsymbol{X}_{i}^{j} - \boldsymbol{X}_{i}^{o_{i}} \right\rVert ^2 \right\} 
\label{keypoints}
\end{align}
where $o_{i}$ is the center of the $i$-th object point cloud.
$\mathbb{P}_{j}^{m_{i}}$ represents the $j$-th point within one of the $k$ points that belongs to the mask $\boldsymbol{m}_{i}$.
$\boldsymbol{X}_{i}^{j}$ and $\boldsymbol{X}_{i}^{o_{i}}$ denote the features of the $j$-th point and the center point of the $i$-th object, respectively.
The keypoints $\boldsymbol{p}_{i}$ are projected onto $\boldsymbol{I}$ as numbered visual markers, represented as $\boldsymbol{I}_{\boldsymbol{p}}$.
This visual prompting approach allows the VLM to bridge the gap between geometric segments and semantic understanding without explicit category labels.
Inspired by~\cite{CoRL2024Huang}, we derive the task-level objectives by taking $\boldsymbol{T}$ and $\boldsymbol{I}_{\boldsymbol{p}}$ as inputs:
\begin{align}
\mathcal{G}_{\mathrm{task}} = \left\{ \mathcal{G}_{1}, \mathcal{G}_{2}, \ldots, \mathcal{G}_{l} \right\}  = \text{VLM}(\boldsymbol{T}, \boldsymbol{I}_{\boldsymbol{p}}, \mathcal{P})
\label{VLM}
\end{align}
where $\mathcal{P}$ is the prompt template \footnote{Please refer to the source code available at \url{https://give-me-scissors.github.io/}}.
$\mathcal{G}_{i}$ is the $i$-th stage sub-goal, taking the form of $L_2$ norm objective function associated with the keypoints.
Meanwhile, VLM determines the grasp and release action stages.
To identify the surgeon's interaction intentions, a hand landmarks detector Mediapipe~\cite{lugaresi2019mediapipe} is applied in task planning to locate the keypoint of the surgeon's hand near the robot's space.
After minimizing the loss function and performing interpolation, a series of desired Cartesian positions are obtained for the dual-arm robot.
Then the desired joint configurations $\boldsymbol{q}_{\mathrm{de}}$ are calculated via inverse kinematics (IK) for the QP framework.


\section{Experiments and Results}

\subsection{Experimental Setup}
The experimental platform consisted of both simulation and real-world components. 
The dual-arm robotic system comprised two Franka Research 3 robotic arms. 
Two industrial computers (MIC-770-V2, Advantech, China) with Intel Core i7-10700 CPUs and 8 GB memory, running Ubuntu 22.04 LTS with the PREEMPT\_RT kernel at 1 kHz, served as the low-level controllers. 
For high-level operations, two separate workstations were employed: the first, equipped with an Intel Core i9-14900KF CPU and an NVIDIA RTX 4090 GPU, was dedicated to high-level task planning and QP optimization; the second workstation was responsible for RGB-D camera data acquisition and real-time perception processing. 
Three Intel RealSense D435i RGB-D cameras provided overlapping views for environmental perception. 
ROS2 served as the middleware to connect all system devices. 
The Flexible Collision Library (FCL)~\cite{pan2012fcl} was utilized to generate the ground truth for minimum distances.
Numerical computations of QP optimization were performed using the SLSQP solver within the SciPy library.

\begin{figure}[htpb] 
\centering
\includegraphics[width=\linewidth]{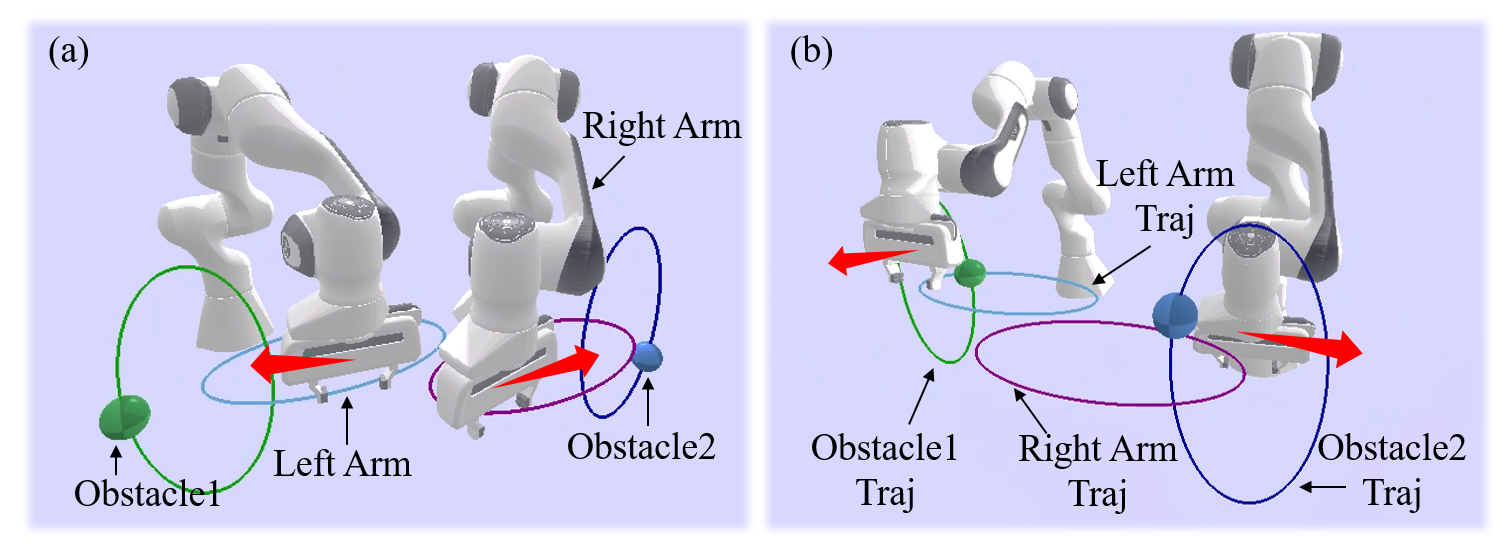}
\caption{
Obstacle and self-collision avoidance process of the dual-arm robot in simulation.
The red arrows indicate the avoidance directions for dual-arm robot.
(a) Self-collision avoidance. 
(b) Obstacle avoidance.
}
\label{SimulationScene}
\end{figure}

\subsection{Collision Avoidance Simulation Experiment}

\begin{figure}[htpb] 
\centering
\includegraphics[width=.85\linewidth]{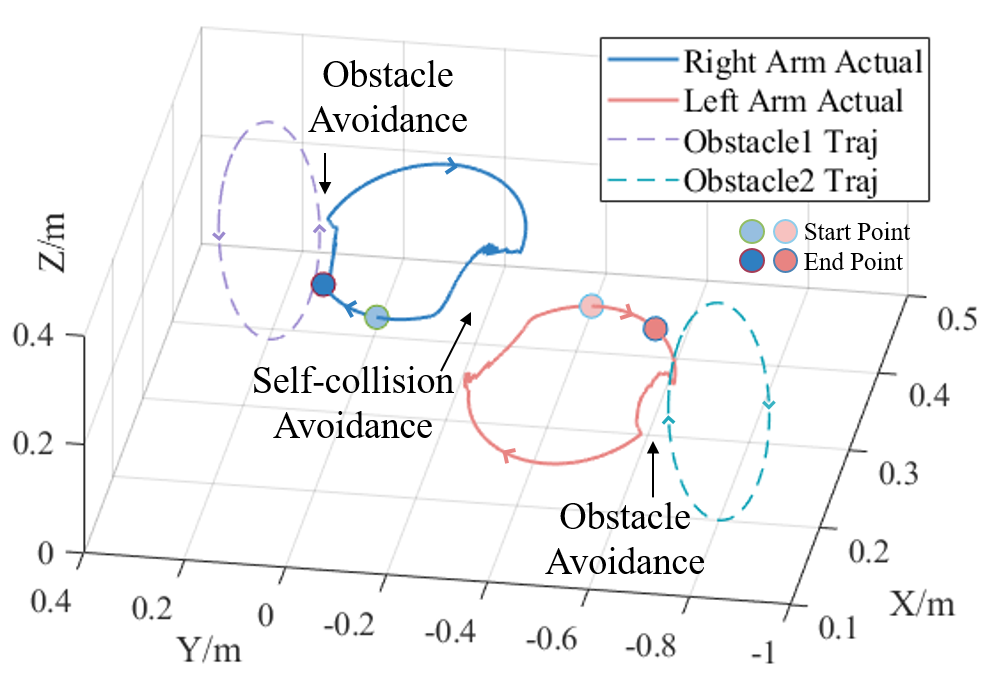}
\caption{
The trajectories of dual-arm robot's end-effector and obstacles during simulation.
The avoidance motion is achieved by our method.
}
\label{SimulationTraj}
\end{figure}

\subsubsection{Implementation Details}

To evaluate the performance of proposed collision avoidance QP framework, simulation experiments of dual-arm robot collision avoidance were conducted.
The robot model was constructed within the PyBullet corresponding to the real robot.
Each robotic arm's end-effector followed an elliptical trajectory (see Fig.~\ref{SimulationScene}(a)), which led to potential self-collision between arms.
Two spheres with a radius of 3 cm (see Fig.~\ref{SimulationScene}(b)) were generated to follow elliptical trajectories near the dual arms, simulating the dynamic obstacles in the environment.
$\hat{d}_{\mathrm{env}}{(\boldsymbol{q}, \mathbb{P})}$ was a MLP-based network consisting of 4 hidden layers with 256 neurons, and the training process was consistent with~\cite{koptev2022neural}.
The parameter in $\hat{d}_{\mathrm{self}}{(\boldsymbol{q})}$ and the training process were the same as in~\cite{luo2025tase}.
In the experiment, the safety distance threshold $\lambda$ and $\mu$ in Eq.~(\ref{QPconstraints}) were adopted as 0.1 m.
$\alpha$ was 5, $\beta$ was 1, $\boldsymbol{Q}$ was an identity matrix. 
The joint configurations, end-effector positions and obstacles positions were recorded during the experiment.
The actual values of the minimum self-distance between dual arms and the minimum distance between robot and obstacles were calculated using FCL. 

\subsubsection{Comparison Methods}

Three state-of-the-art (SOTA) reactive collision avoidance methods~\cite{marangoz2023dawnik, rakita2021collisionik, DingISMCScbf} were compared.
All these methods were extended to encompass both obstacle avoidance and self-collision avoidance.
(1) DawnIK~\cite{marangoz2023dawnik} incorporates collision avoidance as an optimization objective.
$\epsilon_1, \epsilon_2$ = 0.1 (represented in meters).
$\lambda$ and $\mu$ are consistent with Eq.~(\ref{QPconstraints}), and similarly for the following.
(2) CollisionIK~\cite{rakita2021collisionik} encodes distance to collision state using the following cost function.
The scalar values $n,s,c,r$ are the same as in~\cite{rakita2021collisionik}. 
$\epsilon_3$ = 0.1.
(3) CBF-QP~\cite{DingISMCScbf} constructs the collision constraint in QP framework, formulated as:
\begin{align}
\dot{b}( q, \Delta q ) + \gamma (b(q)) \geq 0
\label{CBFQPconstraint}
\end{align}
where $\gamma( r ) = \mathcal{K}r $ is the linear class-$\mathcal{K}$ function.
According to the obstacles collision and self-collision avoidance in our experiment, $b(q)$ can be divided into $b_{\mathrm{env}}(q)$ and $b_{\mathrm{self}}(q)$.
where $b_{\mathrm{env}}(q) = \hat{d}_{\mathrm{env}}{(\boldsymbol{q}, \mathbb{P})} - \lambda$ and $b_{\mathrm{self}}(q) = \hat{d}_{\mathrm{self}}{(\boldsymbol{q})} - \mu$ substitute the first two terms of Eq.~\ref{QPconstraints}.
$\mathcal{K}$ = 0.9.


\begin{table}
\caption{Comparison results between the competing methods and our method}
\label{EvaluationMetricsCompeting}
\begin{center}
\renewcommand{\arraystretch}{1.2}
\setlength{\tabcolsep}{4pt}
\begin{tabular}{c c c c c c c}
\hline
Method & \makecell{ Obs } & \makecell{ Self } & \makecell{ Opt Cost \\ Time (s)} & \makecell{ Mean Pos \\ Error (m)} & \makecell{ Max Accel \\ ($m/s^2$)} \\
\hline
DawnIK~\cite{marangoz2023dawnik} &\ding{56} & \ding{52} & 0.034 & 0.116 & 31.94 \\
CollisionIK~\cite{rakita2021collisionik} & \ding{52} & \ding{52} & 0.038 & 0.136 & 32.10 \\
CBF-QP~\cite{DingISMCScbf} & \ding{52} & \ding{52} & 0.035 & 0.055 & 39.47 \\
\textbf{Ours} & \ding{52} & \ding{52} & \textbf{0.022} & \textbf{0.054} & \textbf{17.77} \\
\hline
\multicolumn{7}{p{215pt}}{'Obs' represents the obstacle avoidance. 'Self' indicates the self-collision avoidance.}
\end{tabular}
\end{center}
\end{table}

\begin{figure}[htpb] 
\centering
\includegraphics[width=\linewidth]{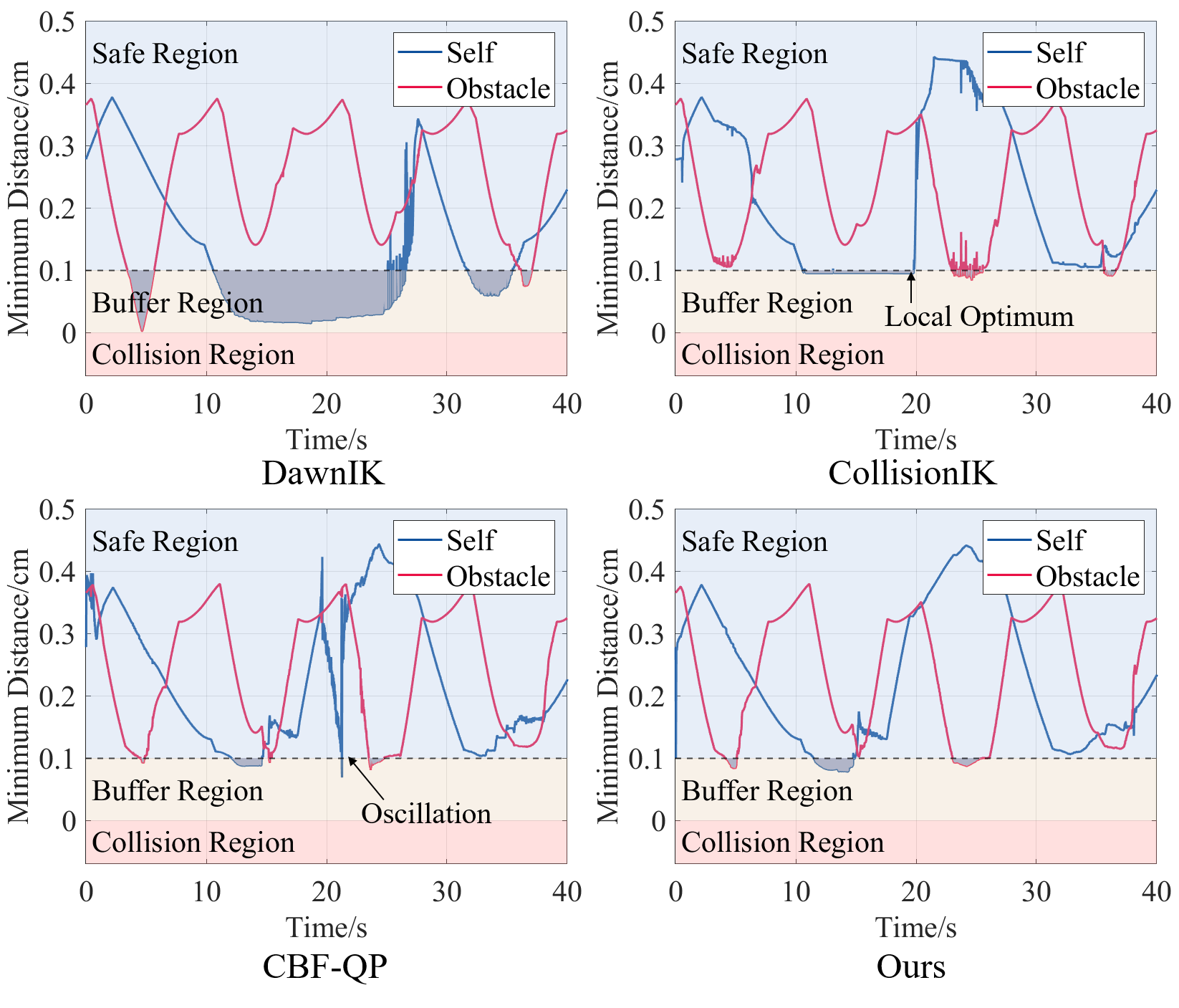}
\caption{
Minimum self-distance between dual arms and minimum distance between right arm and obstacle.
The shadow region represents the portion where the minimum distance falls below the safety distance threshold.
}
\label{SimulationDistance}
\end{figure}


\subsubsection{Results}
The simulation trajectories of the dual-arm robot's end-effectors and the obstacles are illustrated in Fig.~\ref{SimulationTraj}.
The dual-arm robot effectively achieved self-collision avoidance and obstacle avoidance.
Meanwhile, it was able to return to the desired trajectories smoothly after completing the avoidance maneuvers.
The comparative experimental results are shown in Table \ref{EvaluationMetricsCompeting}.
The CollisionIK, CBF-QP and our proposed method successfully accomplished obstacle avoidance, while DawnIK failed (see Fig.~\ref{SimulationDistance}).
All four methods were capable of achieving self-collision avoidance.
Our method has the shortest optimization time, demonstrating better real-time performance.
This is attributed to its rapid convergence during the optimization process.
Additionally, our method exhibits minimum mean position error, indicating that it can closely align with the desired trajectory while ensuring safety.
It also recorded the smallest maximum acceleration among all methods, reflecting smoothness during the avoidance process.
Fig.~\ref{SimulationDistance} illustrates the minimum distance during the simulation among all four methods.
DawnIK encountered an obstacle collision at 4.7s.
CollisionIK got trapped in a local optimum during self-collision avoidance.
CBF-QP exhibited unstable oscillations.
In contrast, our method consistently maintained smooth avoidance motion.

\subsection{Collision Avoidance Real-world Experiment}

\begin{figure*}[thpb]
\centering
\includegraphics[width=\linewidth]{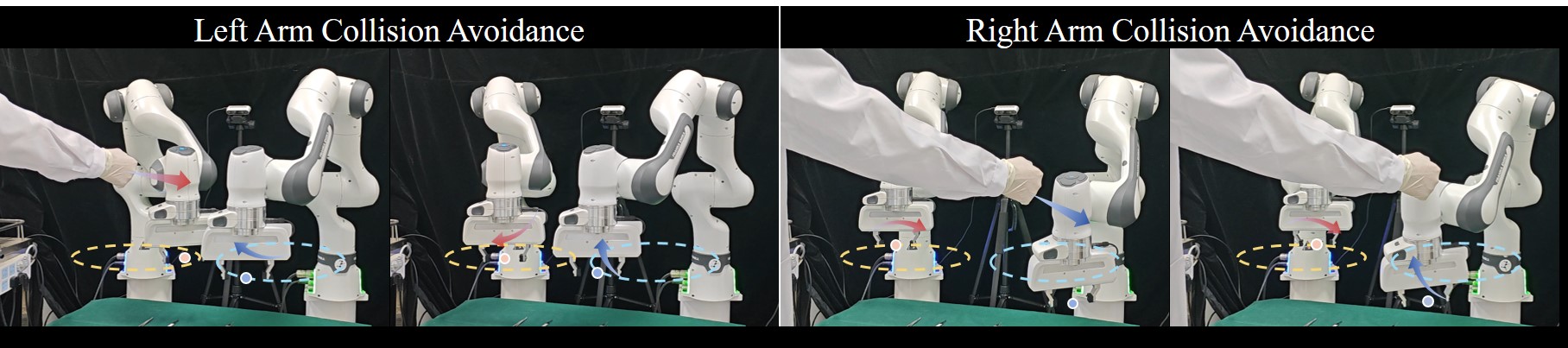}
\caption{
        Dual-arm robot avoids obstacle and self-collision during motion.
        Yellow and blue ellipses represent the desired trajectories of dual-arm robot end-effector.
        Red/blue arrow denotes the motion direction of the left/right arm.
}
\label{RealCollisionScene}
\end{figure*}

\begin{figure*}[thpb]
\centering
\includegraphics[width=\linewidth]{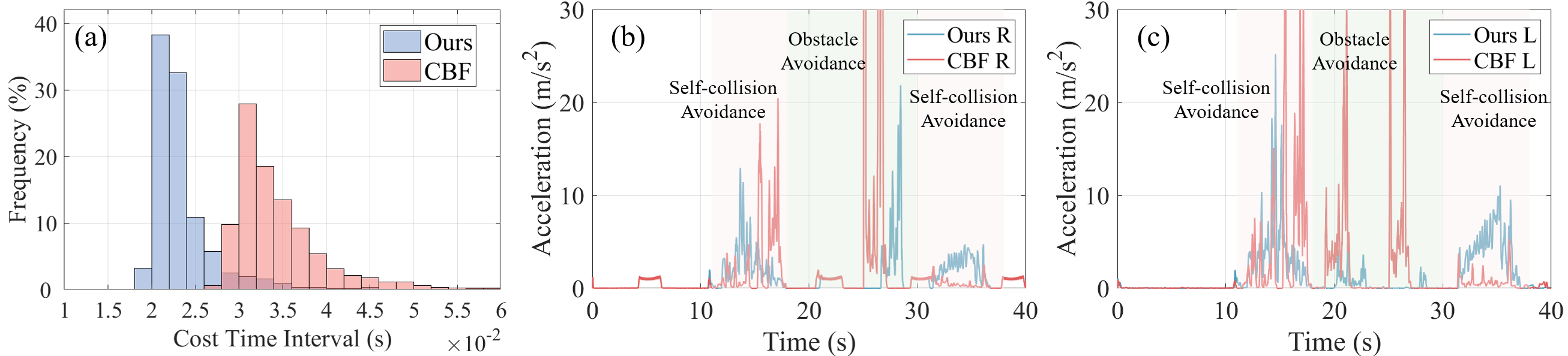}
\caption{
        The results of the collision avoidance real-world experiment.
        (a) Histogram of optimization cost time.
        (b) Right arm end-effector's acceleration.
        (c) Left arm end-effector's acceleration.
}
\label{RealCollisionResults}
\end{figure*}

\subsubsection{Implementation Details}

To verify the safety of the robotics system in complex surgical scenarios, we conducted collision avoidance experiment in real-world settings.
Dual-arm robot's end-effectors followed the same elliptical trajectory as in simulation experiment to verify the self-collision avoidance.
The surgeon approached the dual arms to evaluate the effectiveness of dynamic obstacle avoidance.
Multiple RGB-D cameras were employed to collaboratively capture the point cloud of obstacles near the robot.
The minimum distance between the obstacle and the robot was then calculated using the method proposed in~\ref{MinimumDistancePrediction}.
The safety distance thresholds $\lambda$ and $\mu$ were both adopted as 0.1 m.
Obstacle collision constraint and self-collision constraint were activated when the corresponding minimum distance fell below $\lambda$ or $\mu$.
The radius of the capsules used to approximate the vicinity of the robot was set at 0.25 m.
The CBF-QP mentioned in Eq.~\ref{CBFQPconstraint} was employed as the competing method.
Both the proposed QP framework and CBF-QP operated at a frequency of 40 Hz.
During the experiment, we recorded the optimization cost time and dual-arm end-effector's Cartesian acceleration.
We conducted 10 trials for each method, maintaining the same trajectory.
In each trial, the surgeon approached the left and right arms separately between 18 s and 30 s.
No visual markers were used during the collision avoidance experiments.

\subsubsection{Results}

The experimental process is illustrated in Fig.~\ref{RealCollisionScene}.
When the surgeon approached the dual-arm robot, both arms executed collision avoidance maneuvers while maintaining self-collision avoidance between arms.
The experimental results are presented in Fig.~\ref{RealCollisionResults}.
Compared to the CBF-QP method, the proposed method required less time for optimization (see Fig.~\ref{RealCollisionResults}(a)), demonstrating superior real-time performance.
Fig.~\ref{RealCollisionResults}(b)(c) shows that our method exhibited less jittering during the collision avoidance motion.
This indicates that the nonlinear constraints of proposed QP framework effectively ensure the smoothness of the collision avoidance process.
Both the proposed method and the CBF-QP approach successfully avoided collisions in all trials.
The experimental results demonstrate that the proposed method possesses real-time reactive obstacle avoidance and self-collision avoidance capabilities in dynamic unstructured environments.

\begin{figure*}[thpb]
\centering
\includegraphics[width=.97\linewidth]{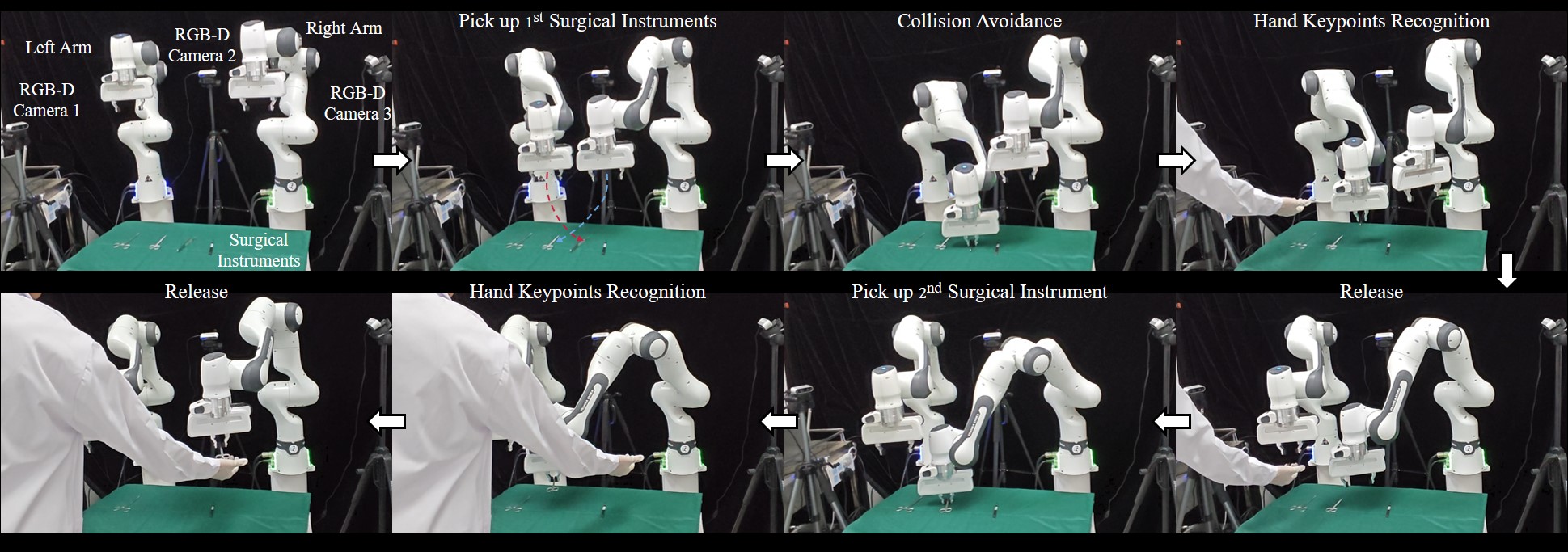}
\caption{
        The experimental process of robot-assisted surgical instrument delivery. 
        The dual-arm robot grasped different surgical instruments in accordance with the surgeon's instructions.
        The surgical instruments were then transferred to the surgeon's hand in sequence.
        The entire process was conducted within the proposed QP framework to prevent collisions.
}
\label{ToolExperiment}
\end{figure*}

\begin{figure}[htpb] 
\centering
\includegraphics[width=.9\linewidth]{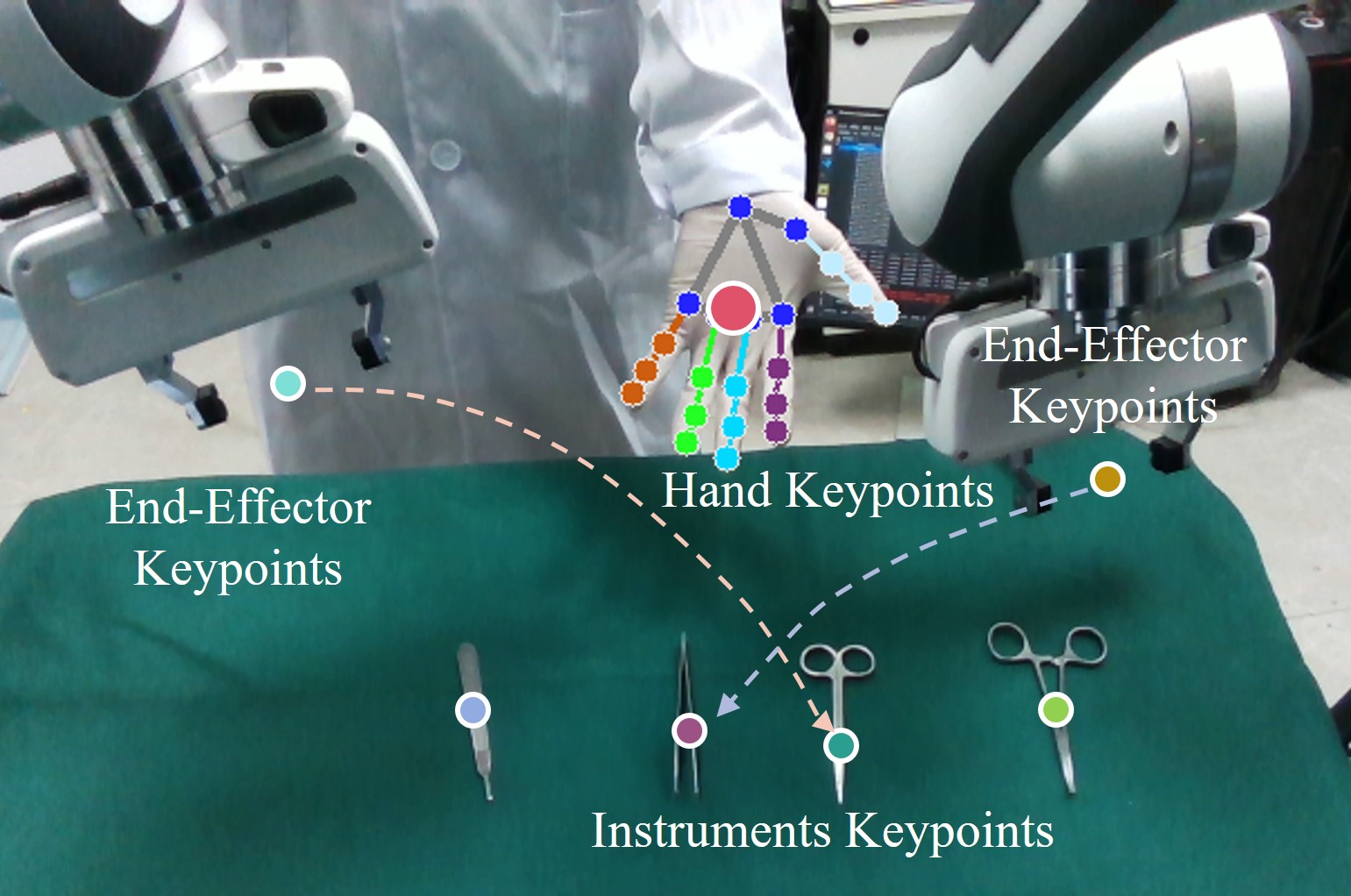}
\caption{
        The keypoints in the experimental process.
        They represent the 3D positions of the corresponding objects.
        The arrows indicate the sub-goals of the robot.
}
\label{KeypointsImage}
\end{figure}

\subsection{Robot-Assisted Instrument Delivery Experiment}

\subsubsection{Implementation Details}

To evaluate the effectiveness of the dual-arm robot in delivering surgical instruments, we designed experiments based on the actual tasks performed by scrub nurses.
Four commonly used surgical instruments (scalpel, tweezer, scissors, and hemostat) were placed on a green sterile drape.
The dual-arm robot was required to autonomously grasp and transfer instruments according to the surgeon's instructions, without relying on predefined pathways.
By implementing the method proposed in~\ref{TaskPlanning}, the sub-goals of the robot were automatically generated by VLM (GPT-4o was applied).
These sub-goals were associated to the keypoints in the environment, illustrated in Fig.~\ref{KeypointsImage}.
The keypoints of the surgical instruments were calculated using the approach in~\ref{TaskPlanning}, while the keypoints of the dual-arm robot end-effectors were obtained through FK.
The average of these landmarks was regarded as the keypoint of the surgeon's hand.
The dual-arm robot was tasked with delivering two surgical instruments simultaneously.
By testing all 6 possible combinations ($ C_4^2 = 6 $) of the four instruments 5 times each, a total of 30 trials were performed. 
This setup ensured that each instrument was tested 15 times in total.
The positions of the instruments were randomly arranged in each trial.
During the experiments, we recorded the number of collisions to evaluate the safety of the robotic system.
We also recorded the number of successful detections, grasps, and deliveries.
Then the overall success rate was calculated.

\subsubsection{Results}

The experimental process for one trial is illustrated in Fig.~\ref{ToolExperiment}.
The dual-arm robot successfully transferred two surgical instruments to the surgeon while avoiding any collisions.
For more details of the experiment, please refer to the supplementary video.
The experimental results are presented in Table~\ref{GraspResults}.
No collisions occurred in any of the trials, validating the reliability and safety of the proposed method.
In terms of detection, no errors were observed with the scalpel and tweezers, while one error occurred with the scissors and hemostat.
This was because the similar shapes of the scissors and hemostat led to misjudgment by the VLM.
The grasping of tweezers was completely successful, while the success rates of the other three instruments were 86.67\%, 85.71\% and 78.57\% respectively.
The reason for the failed grasps was that these instruments were thin and smooth, which made them difficult to grasp on a flat desktop.
The delivery process was entirely successful, except for one instance where the surgeon failed to catch the hemostat.
The average success rate across all trials is 83.33\%, demonstrating the stability and effectiveness of the robotic system.

\begin{table}
\caption{Results of Instrument Delivery Experiment}
\label{GraspResults}
\begin{center}
\renewcommand{\arraystretch}{1.25}
\setlength{\tabcolsep}{4pt}
\begin{tabular}{c c c c c c c c}
\hline
Type & \makecell{ Collision } & \makecell{ Detect } & \makecell{ Grasp } & \makecell{ Delivery } & \makecell{ Success \\ Rate } \\
\hline
Scalpel    & 0 & 15/15   & 13/15   & 13/13   & 86.67\% \\
Tweezer    & 0 & 15/15   & 15/15   & 15/15   & 100.0\% \\
Scissors   & 0 & 14/15   & 12/14   & 12/12   & 80.00\% \\
Hemostat   & 0 & 14/15   & 11/14   & 10/11   & 66.67\% \\
\hline
\end{tabular}
\end{center}
\end{table}

\section{Conclusions and Discussions}

This paper presents a collision-free dual-arm surgical assistive robot for instrument delivery.
The key contribution is the establishment of a robotic system that leverages VLM to intuitively understand surgeon's instructions and autonomously plan movements for grasping and delivering surgical instruments.
The dual-arm robotic system operates in real-time within a QP framework, enabling reactive avoidance of obstacles and self-collision during autonomous motion in the dynamic environment.
Simulations and real-world experiments demonstrate that the proposed robotic system is capable of achieving stable and smooth collision-free motion.
It can deliver the surgical instruments required by surgeons with an average success rate of 83.33\%, indicating the effectiveness.

Compared with robotic scrub nurses in previous studies, the proposed surgical assistive robot eliminates reliance on predefined pathways, significantly enhancing the generalization.
The reactive collision avoidance capabilities of the proposed robotic system make it more suitable for dynamic and unstructured surgical environment.
Nonetheless, several limitations persist.
There is a lack of effective grasping strategies for thin and smooth surgical instruments placed on flat surface.
The motion planning relies on the accuracy of keypoints generation and the object recognition by the VLM, as misjudgments can lead to task failure.
In future work, we will explore utilizing the VLM as a monitor for evaluating sub-goals objectives, leveraging its latent world knowledge to achieve closed-loop correction of task planning.

\addtolength{\textheight}{-12cm}   



\bibliographystyle{IEEEtran}
\bibliography{ICRAref}

\end{document}